\documentclass[acmsmall]{acmart}

\usepackage{graphicx}
\usepackage{eqparbox}
\usepackage{amsthm}
\usepackage{bm}
\usepackage{acronym}
\usepackage{todonotes}
\usepackage{color}
\usepackage{tabularx}
\usepackage{multirow}
\usepackage[ruled,vlined]{algorithm2e}
\usepackage[flushleft]{threeparttable}
\usepackage{makecell}
\SetKwComment{Comment}{/* }{ */}

\usepackage{xcolor}

\newcolumntype{P}[1]{>{\centering\arraybackslash}p{#1}}
\newcolumntype{M}[1]{>{\centering\arraybackslash}m{#1}}

\usepackage{amsmath,bm}
\theoremstyle{plain}

\newtheorem{proposition}{Proposition}

\theoremstyle{definition}
\newtheorem{definition}{Definition}

\newtheorem{remark}{Remark}
\newtheorem*{remark*}{Remark}

\newcommand{\red}{\color{red}}

\newcommand{\iid}{i.i.d.\xspace}

\newcommand{\pth}[1]{\left( #1 \right)}

\newcommand{\sth}[1]{\left\{ #1 \right\}}

\AtBeginDocument{%
  \providecommand\BibTeX{{%
    \normalfont B\kern-0.5em{\scshape i\kern-0.25em b}\kern-0.8em\TeX}}}

\setcopyright{acmcopyright}
\copyrightyear{2023}
\acmYear{2023}
\acmDOI{XXXXXXX.XXXXXXX}

\acmJournal{JACM}
\acmVolume{xx}
\acmNumber{xx}
\acmArticle{xxx}
\acmMonth{8}

\begin{document}

\title[Detecting Abnormal Behavior via Information Sharing]{Towards Safe Autonomy in Hybrid Traffic: Detecting Unpredictable Abnormal Behaviors of Human Drivers via Information Sharing }

\author{Jiangwei Wang}
\email{jiangwei.wang@uconn.edu}
\affiliation{%
  \institution{University of Connecticut}
  \streetaddress{371 Fairfield Way, Department of ECE }
  \city{Storrs}
  \state{CT}
  \country{USA}
  \postcode{06268}
}

\author{Lili Su}
\email{l.su@northeastern.edu}
\affiliation{%
  \institution{Northeastern University}
  \streetaddress{100 Forsyth St, Department of ECE}
  \city{Boston}
  \state{MA}
  \country{USA}
  \postcode{02115}
}
\author{Songyang Han}
\email{songyang.han@uconn.edu}
\affiliation{%
  \institution{University of Connecticut}
  \streetaddress{371 Fairfield Way, Department of CSE }
  \city{Storrs}
  \state{CT}
  \country{USA}
  \postcode{06268}
}

\author{Dongjin Song}
\email{dongjin.song@uconn.edu}
\affiliation{%
  \institution{University of Connecticut}
  \streetaddress{371 Fairfield Way, Department of CSE }
  \city{Storrs}
  \state{CT}
  \country{USA}
  \postcode{06268}
}

\author{Fei Miao}
\email{fei.miao@uconn.edu}
\affiliation{%
  \institution{University of Connecticut}
  \streetaddress{371 Fairfield Way, Department of CSE }
  \city{Storrs}
  \state{CT}
  \country{USA}
  \postcode{06268}
  \thanks{This work was partially supported by NSF Grants 1932250, 1952096, and 2047354 (CAREER).}
}
\renewcommand{\shortauthors}{Wang, Su, Han, Song, and Miao}

\begin{abstract}
Hybrid traffic which involves both autonomous and human-driven vehicles would be the norm of the autonomous vehicles' practice for a while.    
On the one hand, unlike autonomous vehicles, human-driven vehicles could exhibit sudden abnormal behaviors such as unpredictably switching to dangerous driving modes -- putting 
its neighboring vehicles under risks; such undesired mode switching could arise from numbers of human driver factors, including fatigue, drunkenness, distraction, aggressiveness, etc. 
On the other hand, 
modern vehicle-to-vehicle (V2V) communication technologies enable the autonomous vehicles to efficiently and reliably share the scarce run-time information with each other~\cite{dsrc2009dedicated}. In this paper, we propose, to the best of our knowledge, the first efficient algorithm that can 
(1) significantly improve trajectory prediction by effectively fusing the run-time information shared by surrounding autonomous vehicles, and can
(2) accurately and quickly detect abnormal human driving mode switches or abnormal driving behavior with formal assurance without hurting human drivers' privacy. 

To validate our proposed algorithm, we first evaluate our proposed trajectory predictor on NGSIM and Argoverse datasets and show that our proposed predictor outperforms the baseline methods. Then through extensive experiments on SUMO simulator, we show that our proposed algorithm has great detection performance in both highway and urban traffic. The best performance achieves detection rate of $97.3\%$, average detection delay of 1.2s, and 0 false alarm.


\end{abstract}




\maketitle

\section{Introduction}
\label{sec: intro}
\vspace{-2pt}
Despite the rapid development of autonomous vehicles 
in past decades, hybrid traffic which involves both autonomous and human-driven vehicles would be a norm for a long time \cite{bimbraw2015autonomous, veres2011autonomous,huang2016autonomous}.  In this work, we exploit the safety advantages raised by the extended sensing capability of autonomous vehicles through beneficial information sharing. 

Enabling safe autonomy of the autonomous vehicles in the presence of human-driven vehicles is challenging. Human-driven vehicles could exhibit sudden abnormal behaviors such as unpredictably switching to dangerous driving modes. These switches might arise from factors such as fatigue, drunkenness, distraction, and aggressiveness. If not detected in a timely manner, such unannounced switches could quickly put their neighboring vehicles under serious safety threats. To the best of our knowledge, most literature on abnormal driving behavior detection has been focusing on monitoring either behavioral parameters such as eye blinking and yawning \cite{yan2016video, hu2019driving, shahverdy2020driver, lemley2019convolutional,reddy2017real}, or vehicular parameters such as speed variability, steering wheel angle, and steering wheel grip force \cite{li2017online, zhenhai2017driver}, which require placing sensors on vehicle parts like steering wheel, accelerator or brake pedal; see Section \ref{sec: abnormal behavior detection} for details. Unfortunately, human-driven vehicles in the hybrid traffic might not have the required sensor placements to collect the relevant run-time measurements. 
Moreover, such measurements, even if available, are privacy sensitive and may not be shared with other vehicles. When autonomous vehicles does not have the direct measurement data from on-board sensors of the human-driven vehicles, and can only observe human-driven vehicles as part of the driving environment based on autonomous vehicle sensors, it is still challenging to detect abnormal driving behavior. 

On the positive side of hybrid traffic,  modern vehicle-to-vehicle (V2V) communication technologies enable the autonomous vehicles to efficiently and reliably share the scarce run-time information with each other~\cite{dsrc2009dedicated}. Sharing such information can be highly beneficial: the U.S. Department of Transportation (DOT) has estimated that V2V communication 
can address up to 82\% of all crashes in the United States involving unimpaired drivers, potentially saving thousands of lives and billions of dollars~\cite{DSRC_standard}. In addition, navigation and control strategies based on V2V shared information can also improve both traffic efficiency and safety~\cite{han_cdc19}.

\paragraph{\bf Contributions:}
We propose, to the best of our knowledge, the first efficient algorithm that can  
(1) significantly improve trajectory prediction accuracy by effectively fusing the run-time information shared by surrounding autonomous vehicles,
 and can (2)  accurately and quickly detect abnormal human driving mode switches with formal assurance without hurting human drivers' privacy. 
Our algorithm consists of two major components: trajectory prediction component and switch detection component. 
\begin{itemize}
\item On trajectory prediction: We adapt the recently proposed transformer network to the applications of connected and autonomous vehicles (CAVs). 
Moreover, we propose multi-encoder attention mechanism to effectively using the shared information among CAVs. Our model is named as Multi-Encoder Attention based Trajectory Predictor (MEATP).    
We evaluate our proposed MEATP on NGSIM~\cite{I-80} and Argoverse \cite{chang2019argoverse} datasets, the experiment results show that MEATP with information sharing outperforms the well-adopted Long-Short Term Memory (LSTM) \cite{hochreiter1997long} and some transformer \cite{vaswani2017attention} based trajectory predictors. Moreover, information sharing improves the prediction performance of MEATP by $50\%$.
\item On switch detection: We adapt the CuSum algorithm \cite{page1954continuous} and its variant to monitor the patterns of the run-time prediction errors observing that the prediction errors  before and after abnormal driving behaviors can be well captured by two different probability distributions. 
The choice of the CuSum algorithm is motivated by its ease of implementation and strong provably optimality guarantees for formal assurance.   
In this work, we consider two popular scenarios: (1) CAVs have full knowledge of both the pre-change and post-change probability distributions and (2) CAVs have full knowledge of the pre-change distribution only and partial knowledge of the post-change distribution. Note that before the change happens, the driving mode of the human driver is deemed to be normal. Hence the corresponding prediction error distribution can be efficiently computed from existing datasets. Through experiments on SUMO-generated highway and urban traffic datasets, we show that: our proposed algorithm has great detection performance in both highway and urban traffic; it can be generalized to different scenarios where we have full or partial knowledge about post-change distributions; equipped with MEATP with shared information, it is robust to the noises in observations of the surrounding autonomous vehicles. The best performance achieves detection rate of $97.3\%$, average detection delay of 1.2s, and 0 false alarm. 
\end{itemize} 

\section{Related Work}
\label{sec: related work}

\subsection{Trajectory prediction and transformer neural network}

Trajectory prediction is challenging because of the large amount of latent variables involved such as end goals, intentions at each moment, driving styles of human drivers and interactions with other vehicles. There has been extensive progress in vehicle trajectory prediction. Existing literature is divided into three categories: physics-based model, maneuver-based model and interaction-aware model \cite{lefevre2014survey}. 

 Classical dynamic or kinetic model based methods have been designed to predict the future trajectories based on the current state of the vehicles \cite{barth2008will, tan2006dgps, batz2009recognition,  lytrivis2008cooperative, veeraraghavan2006deterministic}. However, they are unable to predict any change in the vehicle motion caused by a particular maneuver or external factors. Deep learning techniques have been widely utilized in maneuver-based model and interaction-aware model, yielding higher accuracy and longer prediction time \cite{sadat2019jointly, li2019conditional}. It has also been shown that the interaction-aware model outperforms in the three classes \cite{lefevre2014survey}. Therefore, deep learning-based interaction-aware trajectory prediction has become the main steam in recent years. Many methods have been proposed, including (but not limit to):   
LSTM combined with convolutional neural network (CNN) to predict the vehicle trajectories on US highways from NGSIM dataset~\cite{deo2018convolutional}, LSTM encoder-decoder to produce the probabilistic future location of the vehicles over occupancy grid map \cite{park2018sequence}, LSTM combined with dynamic geometric graph to predict the vehicle trajectories and road agent behavior~\cite{chandra2020forecasting}. 
In addition, attempts have been made to integrate deep learning with model predictive control to reason about the future behavior of nearby vehicles \cite{cho2019deep}.

A transformer network was originally designed for natural language processing~\cite{vaswani2017attention}, recently it has also been modified to process spatial-temporal data, for traffic flow prediction~\cite{TransformerFlow_21}, pedestrian trajectory prediction~\cite{TransformerTrajectory_21} and vehicle trajectory prediction. An encoder-decoder architecture based on multi-head
attention generates the distribution
of the predicted trajectories for multiple vehicles in parallel \cite{kim2020multi}. Dong et al.  present a dynamic graph attention network to deal with social interactions and predict multi-modal trajectories with probability \cite{dong2021multi}. Messaoud et al. apply multi-head attention by considering a joint
representation of the static scene and surrounding agents to generate future trajectories \cite{messaoud2021trajectory}. There are other works applying attention mechanism to trajectory prediction\cite{bhat2020trajformer, liu2021multimodal}. These works employ attention mechanism to detect vehicles that are more likely to affect the target vehicle's trajectory and pay more attention to them. The temporal dynamics of target and surrounding vehicles' trajectories, however, are not considered.

There are two types of predictions based on the design of the neural networks: (1) One autonomous vehicle (ego vehicle) predicts its surrounding vehicles' trajectories, thus all the surrounding vehicles are target vehicles. In this case, the input contain the past trajectories of the ego vehicle and the target vehicles, the outputs are the predicted trajectories of the target vehicles \cite{kim2020multi, park2018sequence}. This type of prediction, however, doesn't take the trajectories of the vehicles that are around the target vehicle into consideration. As these vehicles interact with the target vehicle, missing their trajectories may result in inaccurate prediction. (2) One autonomous vehicle predicts one target vehicle's trajectory \cite{alahi2016social, deo2018convolutional, chandra2020forecasting,cho2019deep, dong2021multi, messaoud2021trajectory}. In this case, the input contain the past trajectories of the target vehicle and its neighboring vehicles. These works assume that the trajectories of target vehicle and all its neighboring vehicles are available. However this assumption is too strong in real world, one autonomous vehicle won't be able to get all vehicles' data around target vehicle for prediction. Therefore, in this work, we propose an multi-encoder attention based predictor that fuses the shared information among connected autonomous vehicles (CAVs) to predict the human-driven vehicles' trajectories.


\subsection{Information sharing for connected vehicles} 
 
 Connected autonomous vehicles (CAVs) have been proposed and studied for a long time. Information sharing of basic safety messages (BSMs) (velocity, position, heading angle, and yaw rate) is beneficial to autonomous vehicles' learning and control approaches in scenarios such as freeways, intersections, and lane-merging~\cite{han_cdc19, Coordinate_CAV, CV_intersection, ort2018autonomous}. Besides BSMs, environment information captured by vision sensors (such as cameras and Lidars) is also useful to improve autonomous vehicles' decision-making, trajectory planning, and perceptions~\cite{buckman2020generating, miller2020cooperative, kim2015impact}. GPS reports from neighboring
vehicles can be used to identify driving hazards \cite{yang2019app}. It has also been shown that the V2V communications among the autonomous vehicles can improve traffic safety, traffic flow stability and throughput \cite{ye2019evaluating, talebpour2016influence, fagnant2015preparing}.

There have been concerns about the privacy when sharing the data with other vehicles. V2V and V2I communications allow information to be transmitted between vehicles for safety reasons, but they also expose
the vehicle’s movements and geographical location to external networks, from which people can access to locate a vehicle driver \cite{glancy2012privacy}. This is a serious problem with location-based data, as human traces are unique, enabling an adversary to trace movements even with limited side information \cite{gambs2014anonymization, gillespie2016shifting}. Also, access to the interconnected AVs’ wireless network enables public and private agencies to conduct remote surveillance of AV users, which can undermine individual autonomy through psychological manipulation and intimidation \cite{glancy2012privacy}. 
 Schoonmaker \cite{schoonmaker2016proactive} highlights the inadequacies of protecting location-based data based on customer consent, as customers accept the terms and conditions without fully understanding them. Surveys are also being conducted to investigate public attitude towards the privacy concerns brought by sharing the data, there are low confident on how the information should be exchanged between two vehicles \cite{wu2020analysis, al2019readiness}. In the US, the new SPY Car Act gives NHTSA the authority to protect the use of (and access to) driving data in all vehicles manufactured for sale in the US \cite{taeihagh2019governing}. All vehicles must provide owners or lessees the ability to stop the data collection, except for data essential for safety and post-incident investigations, and manufacturers are prohibited from using the collected data for marketing or advertising without consent from the owners or lessees \cite{taeihagh2019governing}.
Therefore, in this work, the hybrid traffic is consisted by human-driven vehicles that cannot communicate or don't willing to communicate, and the CAVs.


\subsection{Abnormal human driver's behavior detection}
\label{sec: abnormal behavior detection}
Existing work on abnormal human driver's behavior detection is mostly restricted to monitoring driver's behavioral parameters or vehicular parameters. 
Specifically, there is a large body of literature  
that uses camera to extract the behavioral parameters such as eye closure ratio, eye blinking, head position, facial expressions, and yawning, to classify whether the driver is in fatigued driving or distracted driving mode \cite{yan2016video, hu2019driving, shahverdy2020driver}. A handful of work monitor   
vehicular parameters such as changing patterns, vehicle speed variability, steering wheel angle, and steering wheel grip force -- assuming sensor placements on vehicle parts like steering wheel, accelerator or brake pedal \cite{li2017online, zhenhai2017driver}. However, due to the privacy sensitive nature of the human drivers images and the detailed sensor measurements, 
none of these methods are applicable to our setting of hybrid traffic. In this work, we focus on a different challenge of detecting abnormal behavior based only on the observations of the CAVs.

\section{Problem Description} 
\label{Problem description}
\vspace{-2pt}
%
%

Unpredictable human driver abnormal behavior is a great obstacle to safe autonomy in hybrid traffic systems. The autonomous vehicles might also have problems during driving. There have been large numbers of work dealing with problems in perception \cite{van2018autonomous}, behavior planning \cite{wei2014behavioral, best2017autonovi}, and control of autonomous vehicles \cite{kuutti2020survey}, these specific problems about autonomous vehicles are out of our scope in this paper. In this work, the challenge we focus on is: how to detect the human driver's abnormal behavior with high accuracy in a short time based on shared sensing information among CAVs, without violence of human driver's privacy. Normal driving behavior can be characterized by well-controlled speed, reasonable headways, mild accelerations, decelerations, and lane changes. In contrast, as human driver's abnormal behaviors are often caused by human factors such as drunkenness, aggressiveness, and fatigue, the resulting abnormal behaviors usually show up with unreasonable operations on vehicles, which provides useful clues for detecting them. First, regarding aggressive driving, the drivers' abnormal behavior can be reflected by observable signs from the perspective of autonomous vehicles, such as accelerating suddenly, violating the speed limit, and making frequent lane changes to surpass the other vehicles \cite{bar2011probabilistic}. Second, considering the drunk and fatigued driving, when the driver is intoxicated by alcohol, or affected by tiredness, he or she is more likely to issue a sudden acceleration or deceleration due to a response delay \cite{hu2017abnormal,dawson1997fatigue}, therefore the velocity control can be poor. From this perspective, when we are trying to predict the trajectory of a human-driven vehicle, the distributions of the prediction errors of normal driving mode and abnormal driving mode could be quite different. 

To address the above challenge, in this work, we propose an algorithm that is composed by trajectory predictor component and driving mode switch detection component to detect the abnormal driving behaviors. Our proposed algorithm takes advantages of information sharing among CAVs to boost both prediction and detection performance. In this section, we first describe the hybrid traffic system including autonomous vehicles and human-driven vehicles, then illustrate the details of information sharing that will be utilized in both trajectory prediction and abnormal behavior detection.

\subsection{Hybrid traffic system description}
\label{system}
\begin{figure}
    \centering
    \includegraphics[width=0.7\textwidth]{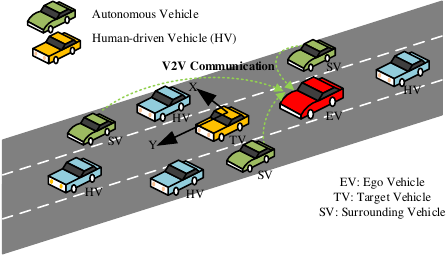} 
    \caption{Hybrid traffic with information sharing.}
    \vspace{-15pt}
    \label{frame}
    \label{fig: illustration}
\end{figure}

The traffic is comprised by autonomous vehicles that are connected via V2V communication and human-driven vehicles. 
For ease of exposition, we refer to the autonomous vehicle that is doing prediction as the ego vehicle (EV), other autonomous vehicles that are within the communication range of the ego vehicle as surrounding vehicles (SV), and the human-driven vehicles is being predicted and as target vehicle (TV). Without loss of generality, we focus on 
one TV. It is easy to see that our algorithm works for the general multiple target vehicles setting under which we can run the algorithm in parallel for different target vehicles. Notably, the human-driven vehicles are not communicating with others. 
The system is illustrated in Fig.\,\ref{fig: illustration}.  

%
%
We define a stationary reference frame based on the TV. 
At time step $t_0$, the origin are set at TV's current location, $X-axis$ is lateral direction, while the $Y-axis$ is longitudinal direction. In the range of $y=-30m$ to $y=30m$, there are in total $N$ neighboring vehicles that interact with the TV, including $N_A$ autonomous vehicles and $N_H$ human-driven vehicles. The range of +30 meters and -30 meters is proposed by Deo et al. \cite{deo2018convolutional} and still widely used in recent works \cite{messaoud2021trajectory, lin2021vehicle}. This setting has been tested and validated on Highway and Urban datasets including NGSIM and nuScenes.

\subsection{Information shared to the ego vehicle}
\label{information}

All surrounding vehicles' observations include point cloud data from Lidar and camera image. With the Lidar-camera fusion techniques \cite{caltagirone2019lidar}, surrounding vehicles can get the locations of the nearby human-driven vehicles. At time $t_0$, For each surrounding vehicle, the information it shares with EV include: 
\begin{itemize}
\item[(1)] its own GPS locations in the past $t_h$ time steps, 
\item[(2)] {\red }locally sensed GPS locations of the nearby human-driven vehicles in the past $t_h$ time steps. 
\end{itemize}
Upon receiving the shared information, EV transfers the GPS locations into the coordinates in current reference frame:  for $i\in[0, N]$ vehicles, coordinates in the past $t_h+1$ time steps: $x_i(t_0-t_h), y_i(t_0-t_h), \dots, x_i(t_0), y_i(t_0) $. 

In real world applications, it's unrealistic for the EV to get accurate GPS locations of all the human-driven vehicles only from the observations of the CAVs. There will be noises in the measurement due to a variety of factors such as  the limitation of on board sensors, the error of object detection and tracking algorithms, etc \cite{yin2021center, fayyad2020deep, cho2019deep}. In this work, we discuss two cases: first is an ideal case, where (1) and (2) are all accurate GPS locations without noise, second is a more realistic case, where (1) are accurate and (2) are noisy. We will elaborate and validate our abnormal behavior detection algorithm against noises in inputs in Section~\ref{noisy} and Section~\ref{sec: experiments} respectively.

It's worth noting that, information sharing among CAVs has more advantages at gathering neighboring vehicles' data compared to existing prediction methods. Multi-agent prediction, usually with multi encoders have been widely used in the existing works \cite{alahi2016social, deo2018convolutional, chandra2020forecasting,cho2019deep, dong2021multi, messaoud2021trajectory}. Each encoder corresponds to one vehicle's data around the target vehicle. However, these existing works assume that the ego vehicle can get all the vehicles' trajectories around the target vehicles, which can be a too strong assumption in some scenarios. When the target vehicle is driving in front of or behind the ego vehicle, shown in Fig. 1, the target vehicle will certainly block part of the ego vehicle's Lidar signal and camera view, therefore the ego vehicle won't be able to get the trajectories of the vehicles that are in front of or behind the target vehicle. The information, however, is vital to predict the target vehicle's trajectory, since the neighboring vehicles' movement can affect the trajectory of the target vehicle. For instance, if the vehicle in front slows down, the target vehicle will also need to slow down to keep a safe distance. In this work, with the shared information among CAVs, other autonomous vehicles can extract, compute, and send the information to the ego vehicle. To the best of our knowledge, we are the first to propose a multi-encoder attention based trajectory predictor that fuses the shared information among the CAVs, and utilize the prediction for our proposed abnormal driving behavior detection Algorithm 1.

\section{Abnormal Behavior Detection Framework}
\label{sec: detection framework}
To make the discussion concrete, we present our framework 
in Section \ref{subsec: human abnormal} -- right after the presentation of our proposed trajectory prediction method (Section \ref{subsec: trajectory prediction}). 
Though our framework is stated w.\,r.\,t.\,the trajectory predictor proposed in Section \ref{subsec: trajectory prediction}  only, it works for more general predictors. Our framework is also generic in its abnormal behavior detection component. In Section \ref{sec: algorithm}, we present three different concrete instantiations to compute the crucial statistic used in the framework. 

\subsection{Multi-encoder attention based trajectory predictor (MEATP)}
\label{subsec: trajectory prediction}
We propose a trajectory predictor, named as {\em multi-encoder attention based trajectory predictor (MEATP)} to fuse the real-time measurement data and utilize the benefit of shared sensing information among CAVs. We specify its inputs and outputs in Section \ref{subsubsec: input output},  introduce its network architecture in Section \ref{subsub: nn architectures}, and present the loss function used in its training in Section \ref{subsubsec: loss function}. 

Transformer based network has been applied in vehicle trajectory prediction\cite{kim2020multi, bhat2020trajformer, liu2021multimodal, messaoud2021trajectory, dong2021multi}. They use attention mechanism to distinguish the vehicles that are more likely to affect the target vehicle's trajectory and give them more attention. In this work, our proposed multi-encoder attention layer, not only finds the more influential vehicles by assigning the corresponding encoder larger weights, but also finds the more important time steps that affects the trajectory of the target vehicles using the multi-head attention layer in the multi-encoder attention. Therefore, our proposed model can find the important features in both spatial and temporal dimensions, whereas the existing works' attention mechanism only focus on spatial dimension.



\subsubsection{Inputs and outputs} 
\label{subsubsec: input output}

The inputs to the MEATP 
include (I.1) the trajectories of neighboring vehicles, 
denoted by $\bm{C}_{i}(t_0)$ for $i\in [1,  N]$, and (I.2) the trajectories of the target vehicles, denoted by $\bm{S}(t_0)$. 
Both (I.1) and (I.2) 
are over a sliding time window $\sth{t_0-t_h, t_0-t_h+1, \cdots, t_0}$, formally,  
\begin{equation}
    \bm{C}_{i}(t_0) \triangleq  [\bm{c}_{i}(t_0-t_h), \cdots, \bm{c}_{i}(t_0-t_h+\ell),  \cdots, \bm{c}_{i}(t_0)]
\end{equation}
with $
\bm{c}_{i}(t_0-t_h+\ell) \triangleq [x_i(t_0-t_h+\ell), y_i(t_0-t_h+\ell)] 
$
being the 2-dimensional 
position of the $i$--th neighboring vehicle at time slot $t_0-t_h+\ell$ for $\ell = 0, \cdots, t_h$,   
and 
\begin{equation}
     \bm{S}(t_0) \triangleq [\bm{s}(t_0-t_h), \dots, \bm{s}(t_0-t_h+\ell), \dots, \bm{s}(t_0)]
\end{equation}
with $\bm{s}(t_0-t_h+\ell) \triangleq [x_0(t_0-t_h+\ell), y_0(t_0-t_h+\ell)]$ being the 2-dimensional vector that records the TV's position at time slot $t_0-t_h+\ell$ for $\ell = 0, \cdots, t_h$. 

The outputs of the MEATP are the distributions of the future trajectory of the TV over time window $\{t_0+1, \cdots,$ $t_0+t_f \}$.  
Assume the predicted future trajectory follows bivariate Gaussian distribution \cite{chandra2019traphic, deo2018convolutional}, notably, this assumption has been widely adopted in the literature \cite{deo2018convolutional,chandra2020forecasting, chandra2019traphic,kim2020multi,messaoud2021trajectory}, tested in numbers of real-world trace datasets, including NGSIM Lyft Level 5, Argoverse Motion Forecasting, and the Apolloscape Trajectory \cite{NGSIM, Chang-2019-121010,ma2019trafficpredict}. the output of the decoder is the bivariate Gaussian parameters at every time step in the future $t_f$ time steps:
\begin{equation}
    \bm{\Omega }\triangleq [\bm{\Omega}(t_0+1),\dots ,\bm{\Omega}(t_0+t_f)], 
\end{equation}
where 
\begin{align*}
    \bm{\Omega}(t) \triangleq [\bm{\mu}(t),\bm{\sigma}(t),\rho(t)] \triangleq [\mu_x(t),\mu_y(t),\sigma_x(t),\sigma_y(t),\rho(t)], ~~~~ \text{for }t \in[t_0+1, t_0+t_f], 
\end{align*}
where $\pth{\mu_x(t), \sigma_x(t)}$ and $\pth{\mu_y(t), \sigma_y(t)}$ are the mean and standard deviation in $x$-axis and $y$-axis, respectively, and $\rho(t)$ is the corresponding correlation-coefficient.

\subsubsection{Network architectures} 
\label{subsub: nn architectures}
\begin{figure}[h]
    \centering
    \includegraphics[width=0.6 \textwidth]{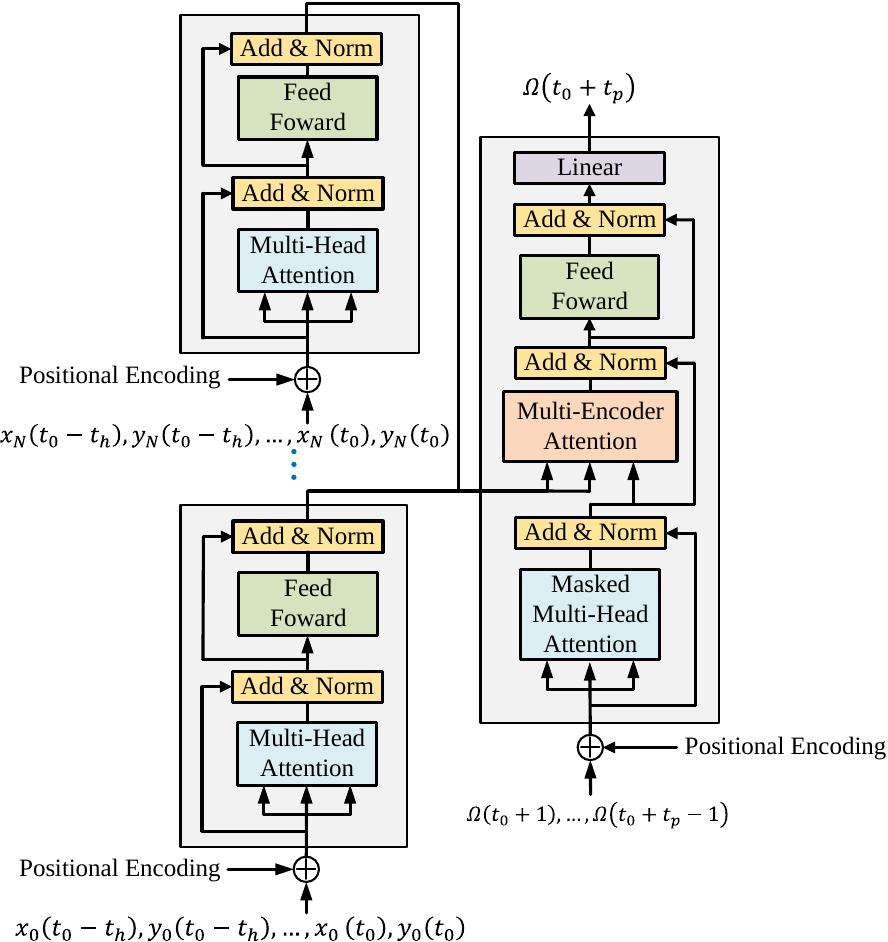}
    \caption{MEATP network architecture. The Multi-Encoder Attention Mechanism in decoder is shown in Fig.~\ref{multiencoderl} }
    \label{encdec}
\end{figure}

 Our network architecture is illustrated in Fig.\,\ref{encdec}. 
To effectively utilize the shared information, 
instead of the single-encoder single-decoder transformer architecture, we use a multi-encoder single-decoder architecture.   
Having multiple encoders enables the EV to learn the spatial and temporal features from the trajectories of the TV and of neighboring vehicles. 
As can be seen from Table \ref{distributions} of our experimental results, compared with CS-LSTM, 
MEATP can more effectively extract the relevant information contained in the shared trajectories, hence can achieve smaller 
mean and variance of the prediction errors.  

Our proposed architecture contains $N+1$ encoders. For ease of exposition, we index these encoders from $0$ to $N$. 
Encoder 0 takes $\bm{S}(t_0)$ -- the target vehicle's trajectory over time window $\sth{t_0-t_h, \cdots, t_0}$ -- as its input. 
Encoder $i$ (for $i=1, \cdots, N$) takes $\bm{C}_{i}(t_0)$ -- the $i$--th vehicle's trajectory over time window $\sth{t_0-t_h, \cdots, t_0}$ -- as its input. The inputs plus the positional encoding are then sent to the next layer as queries $Q$, keys $K$, and values $V$. Positional encoding is just an embedding mechanism to ensure the order of the input.   
Following the seminal work \cite{vaswani2017attention}, each of the encoder consists of  two
sub-layers. The first is a multi-head attention mechanism, and the second is a simple, positionwise
fully connected feed-forward network. We use residual connection around each of
the two sub-layers, followed by layer normalization.  Multi-head attention layer is composed by $h$ heads of scaled dot-product  attentions,  
each of which is expressed as:
\begin{equation}
    \text{Attention}(Q,K,V) = \text{softmax}\pth{\frac{QK^T}{\sqrt{d_k}}}V, 
\end{equation}
where $d_k$ is the dimension of keys. The scaling factor $\frac{1}{\sqrt{d_k}}$is used to prevent the dot product from being numerically too large. Scaled dot-product attention allows the model to pay more attention to one time step in the input that is useful for the prediction. Multi-head attention executes the $h$ scaled dot product attention heads in parallel. 

Attention outputs are concatenated and linearly transformed into the same dimension of Q. Therefore:
\begin{equation}
    \text{Multihead}(Q,K,V) = \text{concat}(\text{head}_1,\dots, \text{head}_h)W^O 
\end{equation}
where  $\text{head}_l= \text{Attention}(QW_l^Q, KW_l^K, VW_l^V)$. With the multi-head attention layer, multiple (instead of just one) time steps of historical data are processed simultaneously.  
Each encoder $i$ passes its  value $V_i$ and key $K_i$.   
to the multi-encoder attention layer of decoder. 

\begin{figure} [h]
    \centering
    \includegraphics[width=0.5\textwidth]{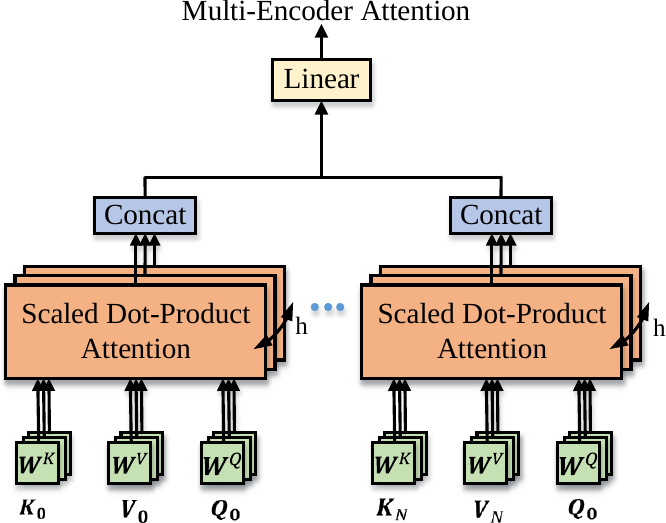}
    \vspace{-5pt}
    \caption{Multi-encoder attention in the decoder}
    \vspace{-10pt}
    \label{multiencoderl}
\end{figure}

Different from the encoders, at time step $t_p$, the decoder takes the outputs in the past $t_p-1$ steps as inputs, and outputs the bivariate Gaussian parameters, which represent the probability distribution of the target vehicle's future coordinates. To be noticed that, the key part in the decoder is the multi-encoder attention. As shown in Fig. \ref{multiencoderl}, the multi-encoder attention  in the decoder is composed by N+1 multi-head attention. Query $Q_0$, which comes from the decoder, interacts with each pair of keys and values $K_i, V_i$ in a multi-head attention. The outputs of the multi-head attention are concatenated and linearly transformed into dimension of $Q_0$. Multi-encoder attention are expressed as:
\begin{equation}
    \text{Multiencoder}(Q_0, K, V) = \text{concat}(M_0,\dots,M_N)W^M
\end{equation}
where $M_i=\text{Multihead}(Q_0,K_i,V_i)$.
The idea of multi-encoder attention comes from the road traffic interaction. Since the target vehicle is continuously interacting with its neighboring vehicles, the past trajectories of the neighboring vehicles, and itself, together influence its future trajectory. By letting query $Q_0$, which corresponds to the target vehicle, interact with $K_i, V_i$, the decoder learns both temporal and spatial information of the neighboring vehicles and target vehicle itself. 

\subsubsection{Loss function} 
\label{subsubsec: loss function}

Since we assume the output at each future time step follows bivariate Gaussian distribution, we train our neural network base on the weighted sum of two losses below:
\begin{equation}
\begin{split}
    &L_1=\sum_{t=t_0+1}^{t_0+t_f}(-\log P(\bm{Z}(t)|\bm{\mu}(t),\bm{\sigma}(t),\rho(t))), \\
    &L_2 = \sum_{t=t_0+1}^{t_0+t_f}||\bm{Z}(t)-\bm{\mu}(t)||,
\end{split}
\end{equation}
where $\bm{Z}(t)=[x_0(t), y_0(t)]$ is the true coordinates of target vehicle at future time step $t$. $L_1$ is the sum (in future $t_f$ time steps) of negative log likelihood of true trajectory given the predicted trajectory distribution. $L_2$ is the $l_2$ norm between true trajectory and predicted trajectory in future $t_f$ time steps. Note that here we choose the mean on $x$ and $y$ axis $\bm{\mu}(t) =[\mu_x(t), \mu_y(t)]$ in output distribution as target vehicle's predicted future trajectory. By minimizing the loss function, we make the predicted trajectory stay close to the future true trajectory. To be noticed that, these two loss functions have been widely used in the exiting literature \cite{deo2018convolutional,chandra2020forecasting, chandra2019traphic,kim2020multi,messaoud2021trajectory}, it is not specifically for transformer networks, all the existing literature uses the true future trajectory in their training phase. In this work, in order to make our abnormal behavior detection algorithm work better, we want the distance between the predicted and true trajectory to be as small as possible. Therefore, the loss function $L2$ are assigned with larger weight. If we denote our proposed predictor as a function $f_W$, with $W$ being the weight parameters, including $W_Q, W_K, W_V$ in scaled dot-product attention layers, $W_O$ in Multihead attention layers, $W_M$ in Multiencoder attention layers, and the weights in the feed forward layers. we have $[\Omega(t_0+1),\dots ,\Omega(t_0+t_f)] = f_W(S(t_0), C_1(t_0), \cdots, C_N(t_0))$. Therefore, by minimizing the loss functions in training phase, we are actually doing gradient decent on the weight parameters $W$ of our proposed model.

 
\subsection{Shared information based abnormal behavior detection algorithm}
\label{subsec: human abnormal}

Recall that the EV is the one that executes the MEATP and monitors whether the TV is in an abnormal driving mode or not. Figuring out which statistic to monitor is highly non-trivial as the driving mode of the TV can switch to an abnormal driving mode at any {\em unknown} time, and the underlying prediction error distributions prior and posterior to the switch time are different. To make the discussion concrete, we use $\gamma$ to denote this unknown switch time. We consider the most challenging scenario wherein no prior information on $\gamma$ is available. 




Recall that $t_0$ is the variable that indicates the current system time; its value increases by 1 as each time goes by.  
At each time, the EV first receives the shared GPS locations from its SVs, transforms them into $\bm{S}(t_0)$ and $\{\bm{C_i}(t_0)\}_{i=1}^N$, and then passes them as inputs to the MEATP to obtain $\bm{\mu}(t_0+1)$. 
Finally, when the true $\bm{Z}(t_0+1)$ is revealed, the EV computes its prediction error  
\begin{equation}
\label{def: prediction error def}
    \bm{e}_{t_0+1}=||\bm{Z}(t_0+1)-\bm{\mu}(t_0+1)||, 
\end{equation}
As $t_0$ increases over time, via the above process, the EV computes a sequence of prediction errors  $\{ \bm{e}_n, n = 1,2, \cdots\}$.

We use $f$ and $g$ to denote the distribution of $\bm{e}_n$ when $n<\gamma$ and $n\ge \gamma$, respectively. 
Clearly, if $\gamma >t_0$, i.e., the TV's driving mode has not switched yet, the TV is currently in a normal mode with
$\bm{e}_n\overset{\iid}{\sim} f$ for $n = 1,2, \cdots, t_0$. 
If $\gamma\le t_0$, i.e., the TV is in the abnormal mode, then $\bm{e}_n \overset{\iid}{\sim} g$ for $n=t_0, \cdots$. 
At any time, the EV is interested in knowing whether a mode switch 
has occurred or not and wants to detect such switch as soon as possible under a given false alarm budget.

\textbf{Therefore we formulate the problem of detecting abnormal human driver behaviors as detecting the change in distributions of the sequence of random prediction errors $\{ \bm{e}_n, n = 1,2, \cdots\}$. } 
The ``distance'' between $f$ and $g$ can be captured by the Kullback-Leibler divergence $D_{\text{KL}}(f,g)$, defined as  
\begin{equation}
 0 < D_{\text{KL}}(f,g) \triangleq \int f(x) \log \pth{\frac{f(x)}{g(x)}} dx <\infty.      
\end{equation}

%

\begin{definition}[Detection algorithm as a stopping time]
\label{def: Detection algorithm as a stopping time}
A {\em stopping time} w.r.t. the sequence of random prediction errors $\{ \bm{e}_n, n = 1,2, \cdots\}$ is a random variable $\tau$ with the property that for each $n$, the event $\{\tau=n\}$ is measurable w.r.t.\,$\sigma(\bm{e}_1, \cdots, \bm{e}_n)$ -- the $\sigma$-algebra generated by $\bm{e}_1, \cdots, \bm{e}_n$. Formally, $\{\tau=n\}\in \sigma(\bm{e}_1, \cdots, \bm{e}_n)$ for each $n$. 
A detection algorithm is a stopping time of random prediction errors that declare the detection of a change \cite{veeravalli2014quickest}. 
\end{definition}

\begin{remark}
A detection algorithm should be a stopping time otherwise it is not implementable due to lack of information, i.e., $\{\tau=n\}\notin \sigma(\bm{e}_1, \cdots, \bm{e}_n)$ for some $n$. When used as a detection algorithm, the event $\tau=n$ is interpreted as {\em ``a distribution change is declared at time $n$''}. 
\end{remark}
%
%


Given the random sequence prediction error $\{ \bm{e}_n, n = 1,2, \cdots\}$, it's hard to tell when the distribution has changed by manual inspection, instead, we compute a statistics $W_n$ given $\{ \bm{e}_n, n = 1,2, \cdots\}$, which will be specified in Section  \ref{sec: algorithm}. 
Once the statistic $W_n$ exceeds some threshold $b$, we declare the change in distributions, i.e., abnormal driving behavior happened. We propose Algorithm \ref{alg: proposed} to illustrate how EV detects whether one TV has turned into abnormal driving mode. Notably, this algorithm can be ran in parallel to detect multiple TVs. Based on the shared information, the EV equipped with trained predictor computes the prediction errors at each time step, the statistic $W_n$ is then updated accordingly.  Once the statistic $W_n$ exceeds the threshold, the EV declares the detection of abnormal behavior. We consider different scenarios regarding the amount of knowledge known on probability distributions $f$ and $g$: (1) We have full knowledge of the pre-change and post-change distributions of $f$ and $g$,  (2) we have full knowledge of pre-change distribution $f$, but we only have partial knowledge about the post-change distribution $g$. 
Assuming full knowledge on pre-change distribution is standard in the literature on quickest change detection \cite{veeravalli2014quickest}. Beside, this assumption can be easily satisfied in our applications:  
 before the change happens, the driving mode of the human driver is deemed to be normal. Hence the corresponding prediction error distribution can be efficiently computed from existing datasets. 
Based on whether we have full knowledge about the post-change distributions or not and, if not, how much we know about it, the statistic $W_n$ can be calculated in CuSum algorithm~\cite{page1954continuous}, MCuSum algorithm \cite{tartakovsky2008quickest, banerjee2014data}, or generalized likelihood ratio test (GLRT)~\cite{lorden1971procedures} approach. We elaborate the details of those algorithms and how to update statistic $W_n$ in Section \ref{sec: algorithm}. 


\begin{algorithm}[h!]
\SetAlgoLined
Initialize $W_0 \gets 0$, $t_0 \gets 0 $, $\mu(t_0) \gets Z(t_0)$\; 
Choose CuSum/ MCuSum/ GLRT algorithm based on the knowledge of the post-change distribution\;
\eIf{Using MCuSum}{
Set threshold $b \gets \text{log} (\frac{M}{\alpha})$ \Comment*[r]{$\alpha$ is the given false alarm budget and $M$ is the number of possible $g$}
}{
Set threshold $b \gets |\text{log}  \alpha|$\; 
}
  \While{true}{
  Receive information shared by the SVs\;
  Update $\bm{S}(t_0)$ and $\{\bm{C}_i(t_0)\}_{i=1}^N$ by incorporating this newly received information \; 
  $\bm{\mu}(t_0+1) \gets f_W(\bm{S}(t_0),\{\bm{C}_i(t_0)\}_{i=1}^N)$  \Comment*[r]{Compute $\bm{\mu}(t_0+1)$ by calling MEATP}
  $\bm{e}_{t_0}\gets||\bm{Z}(t_0)-\bm{\mu}(t_0)||$ \;
  Compute statistic $W_{t_0}$\;
  \If{$W_{t_0} \geq b$}{Declare the detection of abnormal behavior\;
  \bf{Break}\;}
  $t_0 \gets t_0+1$\;
  }
 \caption{Abnormal Behavior Detection based on Shared Information }
 \label{alg: proposed}
\end{algorithm}

\vskip 0.3\baselineskip
\noindent {\bf Minimax quickest change detection (QCD): }
Next we state the performance metrics of the detection algorithms. Observing that in practice the prior knowledge on the switch timing of the driving modes is barely available, we do not impose any distributional assumption of $\gamma$. Instead, we allow $\gamma$ to be an arbitrary and unknown value which can even vary across executions. In other words, we consider minimax QCD. 
The measure of the false alarm is 
the false alarm rate in the literature~\cite{veeravalli2014quickest}:
\begin{equation}
    \text{FAR}(\tau) = \frac{1}{E_\infty(\tau)},
\end{equation}
where $E_\infty$ stands for the expectation measure that the change happens at time $\infty$, i.e., the change doesn't occur. 
To provide strong safety guarantee, we adopt the Lorden's minimax formulation \cite{lorden1971procedures} in which the detection delay is measured via 
\begin{equation}
    \text{WADD}(\tau) = \underset{n \geq 1}{\sup}\:  \text{ess}\: \sup\: E_n[(\tau-n)^+| \bm{e}_1,\cdots, \bm{e}_n], 
\end{equation}
where $(\cdot)^+ = \max\{0, ~ \cdot\}$, $E_n[\cdot]$ is the expectation operator when the change occurs at time $n$, and $\text{ess}\sup (\cdot)$ of a scalar-valued random variable is the sup of its support. 

\subsection{Algorithms to calculate $W_n$ for abnormal behavior detection}
\label{sec: algorithm}

In this section, we illustrate the abnormal behavior detection component. Given whether we have full knowledge about the post-change distributions or not, we update the statistic $W_n$ based on CuSum algorithm, MCuSum algorithm and generalized likelihood ratio test (GLRT) algorithm. We provide details of adopting these algorithms in our problem.

\subsubsection{CuSum algorithm with full knowledge of pre-change and post-change distributions}
We first consider the best case, where we have have full knowledge of the probability distributions $f$ and $g$. For ease of exposition, we assume 
that $f$ and $g$ are gaussian distributions; they can be parameterized by their means and standard deviations as 
$f_\phi$ and $g_\theta$, where each of $\phi$ and $\theta$ is a tuple of mean and standard deviation.
We adapt the CuSum algorithm to detect the abnormal human driver behaviors.  

\noindent \textit{\bf CuSum's algorithm}~\cite{page1954continuous}: 
For some given $b>0$ (chosen later in Proposition \ref{thm: optimality}), 
%
\begin{equation}
\label{eq: CuSum stopping}
    \tau \triangleq  \text{inf} \: \{ n \geq 1 :W_n \geq b \}, 
\end{equation}
where
\begin{align*}
\label{eq: CuSum likelihood}
W_n =  
\begin{cases}
\max\{W_{n-1} +\log L(\bm{e}_{n}), ~ 0\}, &~~~ \text{for } n\ge 1; \\
0, &~~~ \text{for } n = 0. 
\end{cases}
\end{align*}
and $L(\bm{e}_n) = g_\theta(\bm{e}_n)/f_\phi(\bm{e}_n)$ is the likelihood ratio.

\vskip 0.3\baselineskip
%

\vskip 0.3\baselineskip 
The intuitions behind the CuSum algorithm are as follows: 
Upon and after the change point $\gamma$, the prediction error $\bm{e}_n$ follows the distribution $g_\theta$, and $\mathbb{E}_{n\ge \gamma}[\log\pth{g_\theta(\bm{e}_n)/f_\phi(\bm{e}_n)}] = D_{\text{KL}}(g_\theta,f_\phi) >0$. Similarly, before $\gamma$, the prediction error $\bm{e}_n$ follows the distribution $f$, and $\mathbb{E}_{n<\gamma}[\log\pth{g_\theta(\bm{e}_n)/f_\phi(\bm{e}_n)}] = - D_{\text{KL}}(f_\phi,g_\theta) <0$. Thus, with more and more observations on the prediction errors, we expect $\sum_{n: n\ge \gamma} \log\pth{g_\theta(\bm{e}_n)/f_\phi(\bm{e}_n)}$ to bypass the given threshold $b$. In a sense, the $\max\{\cdot, 0\}$ operation in $W_n$ has two effects: (1) it makes a tentative guess on $\gamma$ and (2) it does the first level protection against the random fluctuation in $\log\pth{g_\theta(\bm{e}_n)/f_\phi(\bm{e}_n)}$. A more fine-grained protection against the random fluctuation is controlled by the choice of $b$; the larger the $b$, the more accurate the detection yet the longer delay.

Clearly, there is a trade-off between the false alarm rate $\text{FAR}(\tau)$ and the detection delay; if we can tolerate arbitrary $\text{FAR}(\tau)$, we can achieve 0 detection delay by trivially set $\tau=0$. 
It has been proved in the literature that the CuSum algorithm is optimal \cite{ritov1990decision, moustakides1986optimal}. 
\begin{proposition}~\cite{veeravalli2014quickest} 
\label{thm: optimality}
The CuSum algorithm with $b = |\log \alpha|$ for any given $\alpha\in (0, 1)$ is first order asymptotically optimal. Furthermore, 
the false alarm rate and the average detection of the CuSum algorithm are bounded as follows: 
\begin{align*}
\text{FAR}(\tau) \le \alpha, \, \text{and} ~ ~ \text{WADD}(\tau)  = O\pth{\frac{|\log \alpha|}{D_{\text{KL}}(g, f)}}. 
\end{align*}
\end{proposition}
\subsubsection{MCuSum with unknown parameters}
Recall that we assume that $f_\phi$ and $g_\theta$ are the gaussian distributions. Denote the mean and variance of $f_\phi$ and $g_\theta$ as $(\mu_0,\sigma_0)$ and $(\mu_1, \sigma_1)$, respectively. 
Thus, we have:
\begin{equation}
    \log L(e_n)= \frac{(e_n-\mu_0)^2}{2\sigma_0^2}-\frac{(e_n-\mu_1)^2}{2\sigma_1^2}+ \log (\frac{\sigma_0}{\sigma_1})
\end{equation}
In CuSum's algorithm, we assume that we have full knowledge about the pre-change distribution parameters $\mu_0,\sigma_0$ and post-change distribution parameters $\mu_1, \sigma_1$. However, when the human driver switches to abnormal driving mode, it's more realistic that we cannot know the exact post-change prediction error distributions. But instead, we have some prior knowledge about the post-change distribution and estimate the parameters based on our prior knowledge. In this section, we assume that we have full knowledge of the pre-change distribution $f$, the post-change distribution $g$ with parameter $\theta$:
\begin{equation}
    \theta \in \Theta=\{\theta_1, \theta_2,\dots \theta_M\}
\end{equation}
Therefore, the unknown post-change distribution belongs to a finite set of distributions. In this scenario, We adopt the MCuSum algorithm to detect the abnormal driver's behavior:

\noindent \textit{\bf MCuSum algorithm}~\cite{tartakovsky2008quickest}: 
%
\begin{equation}
\label{eq: MCuSum stopping}
    \tau_{MC} \triangleq  \text{inf} \: \{ n \geq 1 :\max_{j\in \{1, \cdots, M\}}W_n(\theta_j) \geq b \}, 
\end{equation}
where
\begin{align}
\label{eq: MCuSum likelihood}
W_n(\theta_j) =  
\begin{cases}
\left[W_{n-1}(\theta_j) +\log (\frac{g_{\theta_j}(\bm{e}_n)}{f_{\phi}(\bm{e}_n)})\right]^+, &  n\ge 1; \\
0, &  n = 0.
\end{cases}
\end{align}
In Eq.~\eqref{eq: MCuSum stopping}, $[x]^+ \triangleq \max\{x, 0\}$, and, denoting $\phi=\pth{\mu_0, \sigma_0}$ and $\theta_j = \pth{\mu_j, \sigma_j}$, the log likelihood can be written as 
\begin{equation}
        \log\left(\frac{g_{\theta_j}(\bm{e}_n)}{f_{\phi}(\bm{e}_n)}\right) = \frac{(\bm{e}_n-\mu_0)^2}{2\sigma_0^2}-\frac{(\bm{e}_n-\mu_j)^2}{2\sigma_j^2}+\log(\frac{\sigma_0}{\sigma_j}). 
\end{equation}
Thus, to detect a change when the post-change parameter is
unknown, M CuSum algorithms are executed in parallel, one
for each post-change parameter. A change is declared the first
time a change is detected in any one of the CuSum algorithms. In our context, in order to detect the abnormal human driver behavior, we choose several possible means and standard deviations based on the prior information we have about the prediction errors after the abnormal behavior happens. Then we compute the CuSum algorithm in parallel based on each pair of the possible mean and standard deviation, we declare the detection of the abnormal behavior once the statistic evolves above the threshold in any one of the CuSum algorithm.  The asymptotic optimality of the MCuSum algorithm is proved in \cite{tartakovsky2008quickest}.
\begin{proposition}~\cite{tartakovsky2008quickest} 
\label{MCuSum optimality}
The MCuSum algorithm with $b = \log\frac{M}{\alpha}$ for any given $\alpha \geq 0$ is first order asymptotically optimal. Furthermore, 
the false alarm rate and the average detection of the MCuSum algorithm are bounded as follows: 
\begin{align*}
\text{FAR}(\tau_{MC}) \le \alpha, \, \text{and} ~ ~ \text{WADD}^{\theta}(\tau_{MC})  \lesssim \pth{\frac{|\log \alpha|}{D_{\text{KL}}(g, f)}} \text{as}~~ \alpha \rightarrow 0 , \forall \theta \in \Theta 
\end{align*}
\end{proposition}

\subsubsection{Generalized likelihood ratio test with unknown parameters}

In this subsection, we further relax our requirements on the post-change distributions. We assume that we know the values $\mu_0,\sigma_0$, and we only have partial prior information about the distribution $\mu_1, \sigma_1$:
\begin{align*}
    \mu_1-\mu_0 \geq \nu_m, \sigma_1 - \sigma_0 \geq \delta_m
\end{align*}
Then we can no longer use the recursive form to update $W_n$ since we don't know $\mu_1, \sigma_1$. We adopt the generalized likelihood ratio test based approach, we see the sum of log likelihood from $k$ to $n$ as a function of unknown parameter $\theta$,
\begin{equation}
    S_{k}^{n}(\theta)= \sum_{i=k}^{n} \log L(e_i) = \sum_{i=k}^{n} \log(\frac{g_{\theta}(e_i)}{f_{\phi}(e_i)})
\end{equation}
then we have double maximization:
\begin{equation}
    W_n= \max_{1 \leq k \leq n} \sup_{\theta} S_{k}^{n}(\theta)
\end{equation}
Therefore, we can adopt the generalized likelihood ratio test (GLRT) to test the abnormal driving behavior:

\noindent \textit{\bf GLRT algorithm}~\cite{lorden1971procedures}:  
%
\begin{equation}
\label{eq: CuSum stopping}
    \tau \triangleq  \text{inf} \: \{ n \geq 1 :W_n \geq b \}, 
\end{equation}
where
\begin{align*}
\label{eq: CuSum likelihood}
    W_n= \max_{1 \leq k \leq n} \sup_{\theta} S_{k}^{n}(\theta)
\end{align*}

Next we elaborate how to deal with the double maximization in statistic $W_n$. since we know the minimum magnitude change of the parameter $\theta_1$, we have:
\begin{equation}
    W_n= \max_{1 \leq k \leq n} \sup_{
        \theta: \mu_1-\mu_0 \geq \nu_m >0, 
        \sigma_1 - \sigma_0 \geq \delta_m >0 }
     S_{k}^{n}(\theta)
\end{equation}
since 
\begin{align*}
        \frac{(e_n-\mu_0)^2}{2\sigma_0^2}-\frac{(e_n-\mu_1)^2}{2\sigma_1^2}+log(\frac{\sigma_0}{\sigma_1}) > \frac{(e_n-\mu_0)^2}{2\sigma_0^2}-\frac{(e_n-\mu_1)^2}{2\sigma_0^2}+log(\frac{\sigma_0}{\sigma_1})
\end{align*}
we can use the cumulative sum:
\begin{equation}
    S_{k}^{n}=\frac{\mu_1-\mu_0}{\sigma_0^2}\sum_{i=k}^{n}(e_i-\frac{\mu_1+\mu_0}{2}+\log(\frac{\sigma_0}{\sigma_1}))
\end{equation}
let us introduce $\nu = \mu_1 - \mu_0$, $\delta = \sigma_1-\sigma_0$. then $W_n$ can be written as :
\begin{equation}
           W_n= \max_{1 \leq k \leq n} \sup_{
        \theta: \nu \geq \nu_m >0, 
        \delta \geq \delta_m >0 }
     \sum_{i=k}^{n}[\frac{\nu(e_i-\mu_0)}{\sigma_0^2}-\frac{\nu^2}{2\sigma_0^2}+\log(\frac{\sigma_0}{\sigma_0+\delta})]
\end{equation}
in such case, the constrained maximization over $\theta_1$ is:
\begin{equation}
               W_n= \max_{1 \leq k \leq n} 
     \sum_{i=k}^{n}[\frac{\hat{\nu_k}(e_i-\mu_0)}{\sigma_0^2}-\frac{\hat{\nu_k}^2}{2\sigma_0^2}+\log(\frac{\sigma_0}{\sigma_0+\hat{\delta}})]
\end{equation}
where:
\begin{equation}
    \hat{\nu}_k=(\frac{\sum_{i=k}^{n}|e_i-\mu_0|}{n-k+1}-\nu_m)^++\nu_m,  \hat{\delta}=\delta_m
\end{equation}

\subsubsection{Noisy HV location measurements}
\label{noisy}

When SVs share noisy human driven vehicles' GPS locations to the EV, due to the difference in the inputs to our proposed MEATP, the distribution of prediction error changes accordingly. However, the measurement noises have no impacts on $\gamma$. Our algorithm can still detect the change in the distributions even with the noises. For instance, considering MCuSum algorithm, with the noises in the inputs to MEATP, the pre-change distribution of the prediction error changes from $f_\phi$ to $f_{\phi^{\prime}}$, where  $\phi^{\prime}=(\mu_0^{\prime}, \sigma_0^{\prime})$, similarly, post-change distribution changes from $g_{\theta_j}$ to $g_{\theta_j^{\prime}}$,  where  $\theta_j^{\prime}=(\mu_j^{\prime}, \sigma_j^{\prime})$.  
For each $\theta_j^{\prime}$, the log likelihood becomes:
\begin{equation}
        \log\left(\frac{g_{\theta_j^{\prime}}(\bm{e}_n)}{f_{\phi^{\prime}}(\bm{e}_n)}\right) = \frac{(\bm{e}_n-\mu_0^{\prime})^2}{2\pth{\sigma_0^{\prime}}^2}-\frac{(\bm{e}_n-\mu_j^{\prime})^2}{2\pth{\sigma_j^{\prime}}^2}+\log(\frac{\sigma_0^{\prime}}{\sigma_j^{\prime}}). 
\end{equation}

\section{Experiments}
\label{sec: experiments}
\subsection{Trajectory prediction}

\subsubsection{Experiment details}

In this section, we use NGSIM \cite{I-80} and Argoverse 1 motion forecasting\cite{chang2019argoverse} datasets for evaluation. NGSIM dataset consists of trajectories of freeway (US-101 and I-80) traffic sampled at frequency 10Hz over 45 minutes. Argoverse dataset is a curated collection of 324,557 scenarios with each 5 seconds long. Each scenario contains the 2D, birds-eye-view centroid of each tracked object sampled at 10 Hz.  We train our model using Adam optimizer with learning rate of 0.01.  The dimension of the model, also known as number of features, is 16. The number of heads is 8. For the feed forward layer, it contains a linear layer of size (16, 32), a Relu Layer, and another linear layer of size (32,16) in sequential. For NGSIM dataset, We use 3s of historical trajectories to predict the trajectories in future 5s with sample frequency of 5Hz. For Argoverse dataset, the length of historical trajectory is 2s and the length of predicted trajectory is 3s.  In our experiments, we use a server configured with Intel Core i9-10900X processors and four NVIDIA RTX2080Ti GPUs. Our experiments are performed on Python 3.6.0, PyTorch 1.6.0, and CUDA 11.0. For trajectory prediction, we use the standard metrics followed by prior trajectory prediction approaches\cite{chandra2020forecasting, kim2020multi, deo2018convolutional}:
\begin{itemize}
    \item Root mean square error (RMSE) of the predicted trajectories with respect to the true future trajectories.
    \item Average displacement error (ADE): The average RMSE of all the predicted positions and real positions during the prediction window.
    \item Final displacement error (FDE): The RMSE distance between the final predicted positions at the end of the predicted trajectory and the corresponding true location.
\end{itemize}


\subsubsection{Prediction results}

We compare our methods with baselines below:
\begin{itemize}
    \item S-LSTM \cite{alahi2016social}: This model uses a fully connected social pooling layer to deal with the LSTM encoder output, and generates a unimodal distribution for future coordinates. 
    \item CS-LSTM \cite{deo2018convolutional}: This model devises a convolutional social social pooling layer to process the LSTM encoder output, and generates a unimodal distribution for future coordinates. 
    \item MHAPTP \cite{kim2020multi}: This method uses multi-head attention based model for probabilistic vehicle trajectory prediction. 
    \item TraPHic \cite{chandra2019traphic}: This approach uses spatial attention based pooling to perform trajectory prediction of road agents in dense and heterogeneous traffic.
    \item MATF-D and MATF-GAN \cite{zhao2019multi}: A Multi-agent tensor fusion network for contextual trajectory prediction.
    \item CAM and ATTGLOBAL-CAM-NF  \cite{park2020diverse}: A cross-agent attention model with understanding of the scene contexts for trajectory prediction.
\end{itemize}

We evaluate our model in two modes: with and without shared information. When there is no information sharing, we assume that the EV can only get the trajectories of the vehicles that are around it. While with information sharing, we assume that the EV is able to get the TV's neighboring vehicles' historical trajectories and feed them into encoders.

Based on the prediction results in Table~\ref{prediction performance1} and Table~\ref{prediction performance2} we show that: 
\begin{itemize}
        \item Our proposed MEATP with shared information has better prediction performance compared with LSTM based models and transformer based models on NGSIM dataset.  This shows the advantages of our multi-encoder attention mechanism at encoding both spatial and temporal features. Information sharing among the CAVs largely improves the prediction performance. 
    \item Our proposed MEATP with shared information outperforms most baseline methods on Argoverse dataset. It shows slightly worse prediction performance compared with GLOBAL-CAM-NF \cite{park2020diverse}. This is because GLOBAL-CAM-NF encodes the scene context, which requires the bird's eye view of the entire driving environment. However, this information may not be available in real world when predicting the trajectories. Compare to this method, our model doesn't need the scene contexts for prediction.  
\end{itemize}

\begin{table} [h!]
\begin{threeparttable}[h!]
\centering
\caption{\textbf{Prediction results on NGSIM dataset.} Root mean square error (RMSE) over 5 seconds of prediction horizon for models are compared. Our Proposed model MEATP without shared information has better prediction performance than the baselines. Our MEATP with information sharing significantly decreases the RMSE values, at 4s and 5s, the RMSE values are less than half of baselines. }
\begin{tabular}{ M{1.5cm} M{1.5cm}M{1.5cm}M{1.5cm}M{1.5cm}M{1.5cm} M{1.5cm}} 
\hline
Evaluation Metric & Prediction Horizon (s) & S-LSTM & CS-LSTM & MHAPTP & MEATP w/o shared information &  MEATP w shared information\\
\hline
\multirow{6}{4em}{RMSE (m)} & 1 & 0.65 & 0.61 & 0.55 & 0.70 & 0.51 \\ 
& 2  & 1.31 & 1.27 & 0.60 & 1.03 & 0.76 \\ 
& 3  & 2.16 & 2.09 &  1.12 & 1.26 & 0.85 \\ 
& 4  & 3.25 & 3.10 & - & 1.59 & 1.05 \\
& 5  & 4.55 & 4.37 & - & 2.17 & 1.36 \\
\hline
\end{tabular}
    \label{prediction performance1}
\end{threeparttable}

\vspace{-10pt}
\end{table}

\begin{table} [h!]
\begin{threeparttable}[h!]
\centering
\caption{\textbf{Prediction results on Argoverse dataset.} ADE and FDE of models are compared. Our Proposed model MEATP with shared information has better prediction performance than most methods.}
\begin{tabular}{ M{1.25cm}M{1.25cm}M{1.25cm}M{1.25cm}M{1.25cm}M{1.25cm}M{1.25cm}M{1.25cm}} 
\hline
Evaluation Metric & CS-LSTM  & MATF-D & MATF-GAN & TraPHic  & CAM  & GLOBAL-CAM-NF & MEATP w shared information\\
\hline
ADE (m) & 1.39 & 1.35  & 1.26 & 1.04  & 1.13 & 0.80 & 1.13 \\ 
FDE (m) & 2.57 & 2.48 & 2.31 &  3.08  & 2.50 & 1.25 & 2.07 \\ 
\hline
\end{tabular}
    \label{prediction performance2}
\end{threeparttable}

\vspace{-10pt}
\end{table}


\subsection{Human driver abnormal behavior detection}
\subsubsection{Experiment details}

In this section, we use open source simulator Simulation of Urban Mobility (SUMO) \cite{SUMO2018} to generate the dataset. We first construct a highway scenario: a highway with 5 lanes, 1000m length. The traffic volume is 8000 Veh/Hour. Without loss of generality, We then set another urban traffic scenario. We construct a city street with 5 lanes, 1000m length. Two intersections with traffic lights are set 300m and 600m away from the start point of the street. Each branch road contains 4 lanes, with 2 lanes in each direction. Regarding the traffic lights, each of them are set to green light status $80\%$ of one cycle, where one cycle is two minutes. The Traffic volumes are 7000 Veh/Hour on the main road and 200 Veh/Hour on each branch road. The total simulation time is one hour in both scenarios. Notebly, we set the traffic volume according to the real world traffic report \cite{Caltrans}. The vehicles are simulated using Krau{\ss} car following model \cite{krauss1998microscopic}. 1000 vehicles are switched to abnormal mode once they pass a random location on the road. In our experiment, abnormal behaviors include unusual speed control, larger accelerations, decelerations, frequent lane-changing, and small headways, etc. They differ from the normal driving behaviors with altered parameters as shown in Table~\ref{parameters}. Vehicle trajectories are collected at frequency of 10Hz in one hour with label of whether the vehicle has turned into abnormal driving mode. Both highway and urban traffic datasets are divided into training and testing set by the ratio of 7:3. 

\begin{table}[h]
\begin{threeparttable}
    \centering
    \caption{\textbf{Parameters of different driving behavior types}. On longitudinal direction: vehicles in abnormal driving mode have larger acceleration and deceleration abilities, nonusual maximum speed, smaller minimum gaps to leading vehicles, and less perfect driving (sigma denotes the driver imperfection in Krau{\ss} car following model, the larger the more imperfection). On lateral direction: vehicles in abnormal driving mode are less willing to perform cooperative lane changing (larger lcCooperative), they will tend to change lane more frequently to gain high speed (larger lcSpeedGain), and less perfect in lane changing (larger lcSigma). }
    \begin{tabular}{M{4cm} M{4cm} M{4cm}}
    \hline
       Parameters  & Normal driving behavior & Abnormal driving behavior  \\
      \hline
        accel ($m/s^2$) & 2.6 & 7  \\
        decel ($m/s^2$) & 4.5 & 8     \\
        miniGap ($m$) & 2.5 & 1.0     \\
        sigma & 0.1 & 0.8     \\
        maxSpeed (highway) ($m/s$) & 30 & 20 or 45      \\
        maxSpeed (urban) ($m/s$) & 16 & 7 or 25      \\
        speedFactor & 1.0 & 1.2     \\
        lcCooperative & 1.0 & 0.1     \\
        lcSpeedGain & 1.0 & 5.0     \\
        lcSigma & 0.1 & 0.8     \\
        \hline
    \end{tabular}
        \label{parameters}
\end{threeparttable}
    \vspace{-5pt}
\end{table}

\subsubsection{Detection results}

We train our proposed MEATP and the baseline CS-LSTM separately on the normal vehicles' trajectories in SUMO-generated highway and urban traffic datasets. We apply the trained predictors to testing set, then compute the mean and standard deviations of the prediction errors in future 3s. 
The distributions of the prediction errors on highway and urban traffic are shown in table \ref{distributions} and table \ref{distributions on urban} respectively. It can be clearly seen that the means and standard deviations of prediction errors on the abnormal vehicles are larger than those of normal vehicles. This is consistent with our expectation, since the prediction will be less accurate when the abnormal vehicles have more unreasonable behaviors, such as sudden accelerations and decelerations, frequent lane changes, etc. What's more, our proposed MEATP with shared information has smallest mean and standard deviation among three predictors. 

\begin{table}
\begin{threeparttable}
   \caption{\textbf{Distributions of prediction errors on highway dataset.} MEATP with shared information has the smallest mean and the smallest standard deviation. With all the three predictors, the mean and the standard deviation of the prediction errors on abnormal vehicles are larger than those on normal vehicles. }
  \begin{tabular}{M{1.5 cm} M{1.5cm} M{1cm} M{1.3cm} M{1cm} M{1.3cm} M{1cm}M{1.3cm}}
    \hline
    Parameters &
      Prediction horizon (s) &
      \multicolumn{2}{M{2.3cm}}{CS-LSTM} &
      \multicolumn{2}{M{2.3cm}}{MEATP w/o shared information} &
      \multicolumn{2}{M{2.3cm}}{MEATP w shared information}
      \\\hline
    \multirow{4}{*}{Mean (m)} & & Normal & Abnormal & Normal & Abnormal & Normal & Abnormal \\
    & 1 & 0.89 & 1.43 & 0.68 &  1.41 & 0.59 & 1.07 \\
       & 2  & 1.30& 1.96 & 0.85 & 1.72 & 0.73 &  1.54  \\
    &    3  & 2.71 & 3.59 & 1.23 & 2.23 & 0.89  & 1.93 \\
    \hline
    \multirow{3}{*}{SD (m)} & 1 & 0.93 & 1.52 & 1.00 & 1.35 & 0.62 & 1.36 \\
      & 2  & 1.68 & 2.50 & 1.47 & 2.37 & 0.79 &  1.98   \\
    &    3  & 1.94 & 3.73 & 1.57 & 3.56 & 0.88 &  3.17  \\
    \hline
  \end{tabular}
  \label{distributions}
  \end{threeparttable}
\end{table}

\begin{table}
\begin{threeparttable}
\caption{\textbf{Distributions of prediction errors on urban traffic dataset.}}
  \begin{tabular}{M{1.5 cm} M{1.5cm} M{1cm} M{1.3cm} M{1cm} M{1.3cm} M{1cm}M{1.3cm}}
    \hline
    Parameters &
      Prediction horizon (s) &
      \multicolumn{2}{M{2.3cm}}{CS-LSTM} &
      \multicolumn{2}{M{2.3cm}}{MEATP w/o shared information} &
      \multicolumn{2}{M{2.3cm}}{MEATP w shared information}
      \\\hline
    \multirow{4}{*}{Mean (m)} & & Normal & Abnormal & Normal & Abnormal & Normal & Abnormal \\
    & 1 & 1.09 & 1.36 & 0.78 & 1.29 & 0.67 &  0.95 \\
       & 2 & 1.62 & 2.00 & 0.96 & 1.65 &  0.83 &  1.42  \\
    &    3  & 2.78 & 3.43 & 1.32 & 2.40 & 1.01 & 1.83   \\
    \hline
    \multirow{3}{*}{SD (m)} & 1 & 0.88  & 1.73 & 0.89 & 1.31 &  0.69 & 1.35 \\
       & 2  & 1.71 & 2.84 & 1.56 & 2.66 & 1.02 & 1.88      \\
    &    3  & 2.89 & 4.60 & 2.48 &  3.72 & 1.13 & 2.92    \\
    \hline
  \end{tabular}
  \label{distributions on urban}
  \end{threeparttable}
\end{table}

\begin{figure}[h!]
    \centering
    \includegraphics[width=0.5\textwidth]{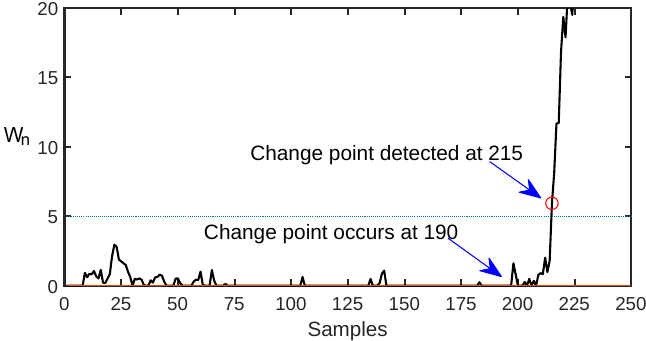}
    \caption{\textbf{Abnormal behavior statistics.} Threshold $b=5$. The abnormal vehicle changes from normal driving mode to aggressive driving mode at time step 190. At time step 215, the statistic $W_n$ exceeds the threshold, thus the detection of abnormal behavior is declared. Detection delay is 25 samples in this case.}
    \label{detected}
\end{figure}
\begin{figure}[h!]
    \centering
    \includegraphics[width=0.5\textwidth]{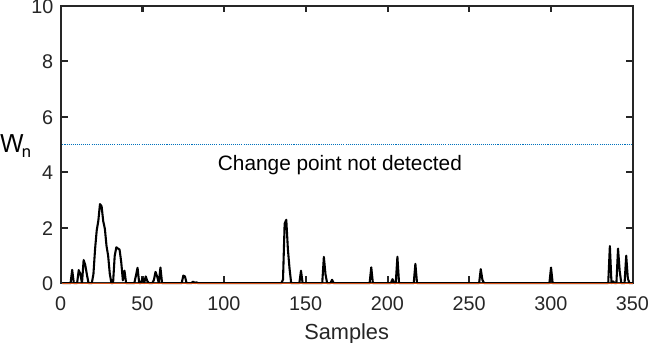}
    \caption{\textbf{Normal behavior statistics.} Threshold $b=5$. Since $W_n$ never exceeds the threshold, the change
    point, namely the abnormal behavior is not detected.}
    \label{notdetected}
\end{figure}

Based on the trained predictors and the probability distributions, we apply proposed algorithm \ref{alg: proposed} to detect the abnormal behaviors. For each target vehicle, at every time step, we use the trained predictor to predict the trajectory in future 5s, then compute the prediction error $\bm{e}_n $ based on the true trajectory and predicted trajectory. After that, $W_n$ is computed based on $\bm{e}_n$ and compared with threshold $b$ to detect the change point.  For each target vehicle, the inference runtime of our detection algorithm equipped with MEATP is 4.8 ms per time step. To be noticed that, depending on whether we have full knowledge of the post-change distribution $g$, we evaluate the algorithm \ref{alg: proposed} based on three different methods: CuSum, MCuSum and GLRT algorithm in highway and urban traffic datasets respectively.  We add four levels of Gaussian noise to the neighboring vehicles' coordinates, with mean and SD being (0.3m, 0.2m), (0.3m, 0.4m), (0.6m, 0.2m), (0.6m, 0.4m) respectively based on the exiting 3D object detection and tracking algorithms \cite{yin2021center, qin2019monogrnet,asvadi20163d}, level 0 means no noise in inputs. The detection results are shown in Table \ref{Cusum} to Table \ref{GLR}. Notably, ADD represents average detection delay in unit of sample. We first summarize our findings based on the results and analyze them separately in the following parts. Our findings:
\begin{itemize}
    \item Our proposed algorithm \ref{alg: proposed}, equipped with well trained predictor, has shown great detection performance in both highway and urban traffic scenarios.
    \item Information sharing among CAVs helps increase the detection rate, lower the false alarms and average detection delay.
    \item our proposed algorithm \ref{alg: proposed} can be generalized to different scenarios: (1) we have full knowledge about the pre-change distribution $f$ and post-change distribution $g$; (2) We have full knowledge about the pre-change distribution $f$ and only partial knowledge about the post-change distribution $g$. In both scenarios, algorithm \ref{alg: proposed} shows remarkable detection performance. When having least information about the post-change distributions, algorithm \ref{alg: proposed} equipped with MEATP with shared information still achieves:  detection rate $91.0\%$, average detection delay 24.9 samples, 16 false alarms in 300 vehicles. 
    \item Equipped with MEATP with shared information, our detection algorithm is robust to the noises in observations of the surrounding autonomous vehicles.
 \end{itemize}
Fig.~\ref{detected} and Fig.~\ref{notdetected} show the statistic evolution given the prediction errors of an abnormal vehicle and a normal vehicle. Notably, the change point, which corresponds to the distribution change, is the point where driver switches from normal driving mode to abnormal driving mode.

\textbf{Detection results based on CuSum algorithm.} In first scenario, we have full knowledge of the pre-change distribution $f$ and post-change distribution $g$. We update statistics $W_n$ based on CuSum algorithm, the detection results are shown in table \ref{Cusum}.  It can be seen that, when having full knowledge of the pre-change and post-change distributions, our proposed algorithm \ref{alg: proposed}  equipped with the well trained predictor can detect the most of the abnormal driving behavior within a short time in both highway and urban traffic scenarios. Even using the baseline CS-LSTM as predictor, our proposed algorithm \ref{alg: proposed} can achieve around $89.0\%$ of detection rate with ADD of 18.7 samples. Compared with the baseline predictor CS-LSTM, our proposed predictor MEATP with shared information outperforms among three predictors, it achieves detection rate of $97.6\%$, 12.3 samples of ADD, and zero false alarms in highway traffic scenario. Because the shared information enables the predictor to learn more about the interactions between target vehicles and its neighboring vehicles, the prediction performance is much better. Based on the small means and standard deviations of the prediction errors, the detection algorithm is more likely to detect the changes in the distributions, less likely to declare false alarm, rendering highest detection rate, smallest false alarms and detection delay. Compared with detection results in highway traffic datasets, the detection results in urban traffic have similar detection rate, yet more false alarms, this may be caused by larger variance in prediction errors in urban traffic dataset. Notably, the detection delay in unit of second is proportional to the sampling rate. The average detection delay of 12.3 samples with 10 Hz sampling rate rendering the detection delay to be around 1.2s. If we want to detect the abnormal human driver's behavior as soon as possible, we can increase the sampling rate. Thus, our proposed abnormal behavior detection algorithm is generic, the detection delay can be improved by improving the hardware sampling capabilities.

By analysis on CuSum algorithm based detection results, we show that:
\begin{itemize}
    \item When having full knowledge of pre-change and post-change probability distributions, our proposed detection algorithm \ref{alg: proposed}, equipped with well trained predictor, is able to detect most of the abnormal behaviors in short time with small amount of false alarms.
    \item Our algorithm \ref{alg: proposed} has remarkable detection performance in both highway and urban traffic scenarios. In highway traffic, the best detection performance is:  0 false alarms, detection rate of $97.6\%$, and ADD 12.3 samples. in urban traffic, the best detection performance is: detection rate $96.3\%$, ADD 18.4 samples , and 6 false alarms in 300 vehicles.
    \item The shared information among autonomous vehicles increases the detection rate and lowers the detection delay, while maintains least false alarms.   
\end{itemize}

\begin{table} [h!]
    \centering
\begin{threeparttable}
   \caption{Detection Results on SUMO dataset based on CuSum}
  \begin{tabular}{ M{1.2cm} M{1.8cm} M{0.6cm}M{0.6cm}M{0.6cm}M{0.6cm}M{0.6cm}M{0.6cm}M{0.6cm}M{0.6cm}M{0.6cm}M{0.6cm}  }
    \hline
   \multirow{2}{5em}{Models} & 
    \multirow{2}{5em}{Parameters} &
    \multicolumn{5}{M{3cm}} {highway traffic \break noise level} &
     \multicolumn{5}{M{3cm}} {urban traffic  \break  noise level}\\
     && 0 & 1 & 2 & 3 & 4  & 0 & 1 & 2 & 3 & 4 \\
      \hline
    \multirow{4}{4em}{CS-LSTM} 
    & detected  & 267 & 264 & 263 & 239 & 226 & 271 & 262 & 258 & 225 & 223  \\
       & false alarm   & 7  & 16 &  26 & 31 & 59 & 20 & 22 & 35  & 59 & 67  \\
    &    ADD    & 18.7 & 31.3 & 30.4 & 28.7 & 25.5 & 30.1 & 25.1 & 32.6 & 33.4 & 31.7  \\
     & detection rate  & $89.0\%$  & $88.0\%$ & $87.7\%$ & $79.7\%$ & $75.3\%$  &  $90.3\%$  & $87.3\%$ & $86.0\%$ & $75.0\%$ &  $74.3\%$ \\
    \hline
    \multirow{3}{4em}{MEATP w/o shared information}
      & detected & 282  &  278  &  270  &  261 &  250  & 277 & 273 &269   &  270 &  259   \\
      & false alarm  & 7  & 16 & 21 &  33 & 42  &  13 & 17 & 25 & 28 &     35   \\
    & ADD & 22.0 & 25.4 & 24.3 & 25.5 & 30.1  & 25.5 & 30.3 & 28.2 & 27.3 & 27.5      \\
     & detection rate  & $94.0\%$ &  $92.7\%$ & $90.0\%$ & $87.0\%$ & $83.3\%$& $92.3\%$ &$89.6\%$ & $91.0\%$ &$90.0\%$ & $86.3\%$ \\
    \hline
        \multirow{3}{4em}{MEATP w shared information} & detected &  293  & 291  & 290  &  288 & 285  &  289    &  284 & 286 & 286 & 283  \\
      & false alarm  & 0 & 1 & 1 & 2 & 4 & 6 & 10 & 8 & 8 & 11   \\
    &    ADD  &  12.3 & 18.3 & 17.1 & 19.6 & 23.2 & 18.4  & 13.6 & 14.8& 15.2  & 18.7  \\
         & detection rate  &  $97.6\%$ & $97.0\%$ & $96.7\%$ & $96.0\%$ & $95.0\%$ &   $96.3\%$   & $94.7\%$ &  $95.3\%$ &  $95.3\%$ & $94.3\%$ \\
    \hline
  \end{tabular}
  \label{Cusum}
  \end{threeparttable}
\end{table}

\begin{table} [h!]
    \centering
\begin{threeparttable}
   \caption{Detection Results on SUMO dataset based on MCuSum}
  \begin{tabular}{ M{1.2cm} M{1.8cm} M{0.6cm}M{0.6cm}M{0.6cm}M{0.6cm}M{0.6cm}M{0.6cm}M{0.6cm}M{0.6cm}M{0.6cm}M{0.6cm}  }
    \hline
   \multirow{2}{5em}{Models} & 
    \multirow{2}{5em}{Parameters} &
    \multicolumn{5}{M{3cm}} {highway traffic \break noise level} &
     \multicolumn{5}{M{3cm}} {urban traffic  \break  noise level}\\
    &  & 0 & 1 & 2 & 3 & 4  & 0 & 1 & 2 & 3 & 4 \\
      \hline
    \multirow{4}{4em}{CS-LSTM} 
    & detected & 263 & 258 & 241 &  225 & 213  & 265 & 260 & 224 &  219 & 207  \\
       & false alarm  & 18 &  32 & 42 & 49 & 60   & 17 &  29 & 51 & 65 & 71   \\
    &    ADD  & 37.3 & 24.0 & 16.7 & 14.5 & 11.7 & 12.5 & 15.9 & 16.8 & 11.6 & 12.8  \\
     & detection rate  &  $87.7\%$ & $86.0\%$ & $80.3\%$ & $75.0\%$ & $71.0\%$  &  $88.3\%$ & $86.7\%$ & $74.7\%$ & $73.0\%$ & $69.0\%$   \\
    \hline
    \multirow{3}{4em}{MEATP w/o shared information}
      & detected & 271  &  267  &  262 &  255 &   253 & 263  &  257  &  253 &  251 &   228\\
      & false alarm  & 12 & 23 & 23 &  26 & 31  & 14 & 21 & 25 &  27 & 57    \\
    & ADD  & 33.9 & 27.9 & 30.2 & 28.6 & 27.3  & 20.7 & 23.7 & 27.2 & 27.6 & 25.7    \\
     & detection rate  &  $90.3\%$ & $89.0\%$ & $87.3\%$ & $85.0\%$ & $84.3\%$  &  $87.6\%$ & $85.6\%$ & $84.3\%$ & $83.7\%$ & $76.0\%$   \\
    \hline
        \multirow{3}{4em}{MEATP w shared information} & detected & 291  &  288 & 287  & 284 & 283 & 287  &  283 & 283  & 281 & 278  \\
      & false alarm  & 4 & 6 & 8 & 10 & 10    & 6 & 10 & 11 & 11 & 15   \\
    &    ADD  &  14.9 & 18.2 & 23.7 & 22.5 & 20.1   &  16.0 & 19.8 & 18.6 & 18.1 & 19.5  \\
         & detection rate  &  $97.0\%$ & $96.0\%$ & $95.7\%$ & $94.7\%$ & $94.3\%$  &  $95.7\%$ & $94.3\%$ & $94.3\%$ & $93.6\%$ & $92.6\%$   \\
    \hline
  \end{tabular}
  \label{MCuSum}
  \end{threeparttable}
\end{table}

\begin{table} [h!]
    \centering
\begin{threeparttable}
   \caption{Detection Results on SUMO urban dataset based on GLRT}
  \begin{tabular}{ M{1.2cm} M{1.8cm} M{0.6cm}M{0.6cm}M{0.6cm}M{0.6cm}M{0.6cm}M{0.6cm}M{0.6cm}M{0.6cm}M{0.6cm}M{0.6cm}  }
    \hline
   \multirow{2}{5em}{Models} & 
    \multirow{2}{5em}{Parameters} &
    \multicolumn{5}{M{3cm}} {highway traffic \break noise level} &
     \multicolumn{5}{M{3cm}} {urban traffic  \break  noise level}\\
    &  & 0 & 1 & 2 & 3 & 4  & 0 & 1 & 2 & 3 & 4 \\
      \hline
    \multirow{4}{4em}{CS-LSTM} 
    & detected & 252 & 233 & 234 & 220 & 201 & 243 & 229 & 231 & 218  &   195  \\
       & false alarm  & 24 & 32 &36  & 56 & 74 &  26  & 32  &  44  & 75 &  75  \\
    &    ADD   & 27.1 & 27.5 & 23.7 & 24.5 & 26.3 & 24.3 & 29.0 & 28.7 & 28.2  & 29.3    \\
     & detection rate  & $84.0\%$  & $77.7\%$ & $78.0\%$ & $73.3\%$  & $67.0\%$   &  $81.0\%$ & $76.3\%$ & $77.0\%$ & $72.7\%$ & $65.0\%$   \\
    \hline
    \multirow{3}{4em}{MEATP w/o shared information}
      & detected & 271 & 259 &  260  & 258  &  248 & 268   & 254 & 253   & 250  &  245   \\
      & false alarm  & 18 &23 & 27 & 27 & 38  & 25 & 32 & 34 & 37 & 50       \\
    & ADD & 22.5 & 25.7& 20.1 & 22.6 &  25.0  &  31.4 & 33.2 & 31.7 & 32.1 &  33.5    \\
     & detection rate  &  $90.3\%$ & $86.3\%$   & $86.7\%$  &  $86.0\%$  & $82.7\%$  & $89.3\%$  & $84.7\%$ & $84.3\%$  & $83.3\%$ & $81.7\%$   \\
    \hline
        \multirow{3}{4em}{MEATP w shared information} & detected &  287 & 286 &283 & 280   &  279 & 283 & 283 &  280 &  278  &  273     \\
      & false alarm  & 5 & 6 & 7 & 11 & 13 &  10 &  11    & 13 & 14  &  16     \\
    &    ADD  &  21.4 & 20.1 & 21.8 &23.3 & 23.9 & 23.0  &  20.4   &  21.0 &  20.2 & 24.9      \\
         & detection rate  &  $95.7\%$ & $95.3\%$ & $94.3\%$ &  $93.3\%$ & $93.0\%$  & $94.3\%$ &  $94.3\%$ &  $93.3\%$  & $92.7\%$ &  $91.0\%$   \\
    \hline
  \end{tabular}
  \label{GLR}
  \end{threeparttable}
\end{table}

\textbf{Detection results based on MCuSum algorithm}. In second scenario, we have full knowledge of the pre-change distribution $f$,  the unknown post-change distribution $g$ with parameter $\theta$ belongs to a finite set of distributions: 
$
    \theta \in \Theta=\{\theta_1, \theta_2,\dots \theta_M\}
$. Regarding the true post-change parameters $\mu_1, \sigma_1$, we add two small magnitude (less than 0.3) gaussian noise to each of them separately, meaning $M=4$ in this case. The detection results based on MCuSum algorithm are shown in table \ref{MCuSum}. The overall detection performance are slightly worse than the first scenario where we have full knowledge of distributions $f$ and $g$. This is consistent with our expectations. Since we only have partial knowledge about the post-change distributions $g$, we can only use the possible set $\theta \in \Theta=\{\theta_1, \theta_2,\dots \theta_M\}$. We perform $M$ CuSum algorithms in parallel,  at every time step, we choose the maximum $W_n(\theta)$ to see whether it exceeds the threshold. 

By analysis on MCuSum algorithm based detection results, it shows that:
\begin{itemize}
    \item There is a trade off between the knowledge we have about post-change distributions and abnormal driving behavior detection performance.    
    \item When we have full knowledge about pre-change distribution $f$, and the unknown post-change distribution $g$ belongs to a finite set, our proposed algorithm 1 still has remarkable detection performance in both highway and urban traffic scenario. Best detection performance based on MEATP with shared information In highway traffic: detection rate detection rate $97.0\%$, ADD 14.9 samples, 4 false alarms in 300 vehicles; in urban traffic: detection rate $95.7\%$, ADD 16 samples , 6 false alarms in 300 vehicles.
    \end{itemize}

\textbf{Detection results based on GLRT algorithm}. In the third scenario, we have full knowledge of the pre-change distribution $f$ with parameter $\phi( \mu_0, \sigma_0)$, for the unknown post-change distribution $g$ with parameter $\theta (\mu_1, \sigma_1)$, we only know the minimum magnitude change compared with pre-change parameters: $
    \mu_1-\mu_0 \geq \nu_m, \sigma_1 - \sigma_0 \geq \delta_m
$. The detection results based on GLRT algorithm are shown in table \ref{GLR}. As can be seen, since we have even less information about the post-change distributions, the detection performance slightly drop compared with second scenario. 

By analysis on GLRT algorithm based detection results, we show that:
\begin{itemize}
    \item When we have full knowledge of the pre-change distribution parameter $\phi$, and we only know that the minimum magnitude difference between post-change parameter $\theta$ and pre-change parameter $\phi$, our proposed algorithm \ref{alg: proposed} still shows great detection performance. MEATP with shared information in highway traffic: detection rate $95.7\%$, ADD 21.40 samples, 5 false alarms in 300 vehicles; in urban traffic: detection rate $94.3\%$, ADD 23.0 samples, 10 false alarms in 300 vehicles. 
\end{itemize}


\section{Conclusion}


This paper proposes abnormal human driving behavior detection algorithm for CAVs based on shared sensing information in hybrid traffic system. This work, to the best of our knowledge, is the first efficient algorithm that can accurately and quickly detect abnormal human driving mode switches based on CAVs sensing data without using in-vehicle sensing data that may hurt human-driver privacy. We first propose a multi-encoder attention based interaction-aware trajectory prediction model called MEATP. Based on the predictor MEATP, we further develop an abnormal behavior detection method. Through extensive experiments on both public dataset and simulator, We show that (1) our proposed MEATP predictor outperforms the baselines; (2) our proposed algorithm detects abnormal behaviors with remarkable high accuracy (the best performance achieves detection rate of $97.3\%$) and low detection delay;  (3) shared information boosts the performance of both trajectory prediction and abnormal behavior detection.



\bibliography{sample-base}

\appendix

\end{document}